\documentclass[letterpaper]{article}
\usepackage{aaai2027}
\nocopyright
\usepackage[hyphens]{url}
\usepackage{graphicx}
\usepackage{natbib}
\usepackage{caption}
\usepackage{amsmath}
\usepackage{amssymb}
\usepackage{booktabs}
\usepackage{colortbl}
\usepackage{algorithm}
\usepackage{algpseudocode}
\usepackage{placeins}
\definecolor{tablegroupgray}{gray}{0.94}
\definecolor{tablefullgray}{gray}{0.88}

\title{ChainVLA: Chaining Vision-Language-Action Queries through a Unified Execution State for Long-Horizon Manipulation}
\author{
    Yuzhi Huang\textsuperscript{\rm 1}\equalcontrib,
    Weijue Bu\textsuperscript{\rm 2}\equalcontrib,
    Ziyi Xiong\textsuperscript{\rm 3},
    Jie Wu\textsuperscript{\rm 1},
    Fanding Huang\textsuperscript{\rm 1},
    Jingyan Jiang\textsuperscript{\rm 3}\thanks{Corresponding authors.},
    Zhi Wang\textsuperscript{\rm 1}\footnotemark[2]
}
\affiliations{
    \textsuperscript{\rm 1}Shenzhen International Graduate School, Tsinghua University\\
    \textsuperscript{\rm 2}China University of Mining and Technology\\
    \textsuperscript{\rm 3}Shenzhen Technology University
}

\begin{document}

\maketitle

\begin{abstract}
Humans perform long-horizon manipulation by retaining knowledge of what earlier actions have established while continuously adapting the motion underway.
By contrast, action-chunked vision-language-action (VLA) policies repeatedly replan from the current input at each query.
Existing methods preserve either long-term task evidence through memory or short-term motion through action reuse and ensembling, leaving the cross-query handoff incomplete.
We introduce ChainVLA, a 1.2B-parameter VLA policy that chains successive queries through a joint and revisable execution state.
Progress Context combines a recurrent Working State with sparse event memory to carry observation-derived task progress, while Motion Tail feeds the preceding prediction's unexecuted continuation into state construction and action generation.
Together, the two components condition a decoder that regenerates each action horizon under the latest observation, allowing the carried state to guide the next prediction without fixing it.
ChainVLA reaches 62.8\% average success on RMBench and 98.8\% across four LIBERO suites, while removing Motion Tail or Progress Context reduces RMBench success to 11.2\% and 3.0\%, respectively.
These asymmetric ablations are consistent with motion continuity helping preserve the observation stream from which task progress is inferred.
\end{abstract}

\begin{links}
    \link{Project page}{https://muqy1818.github.io/chainvla-web/}
\end{links}

\section{Introduction}

\begin{figure*}[t]
    \centering
    \includegraphics[width=0.93\textwidth,pagebox=cropbox]{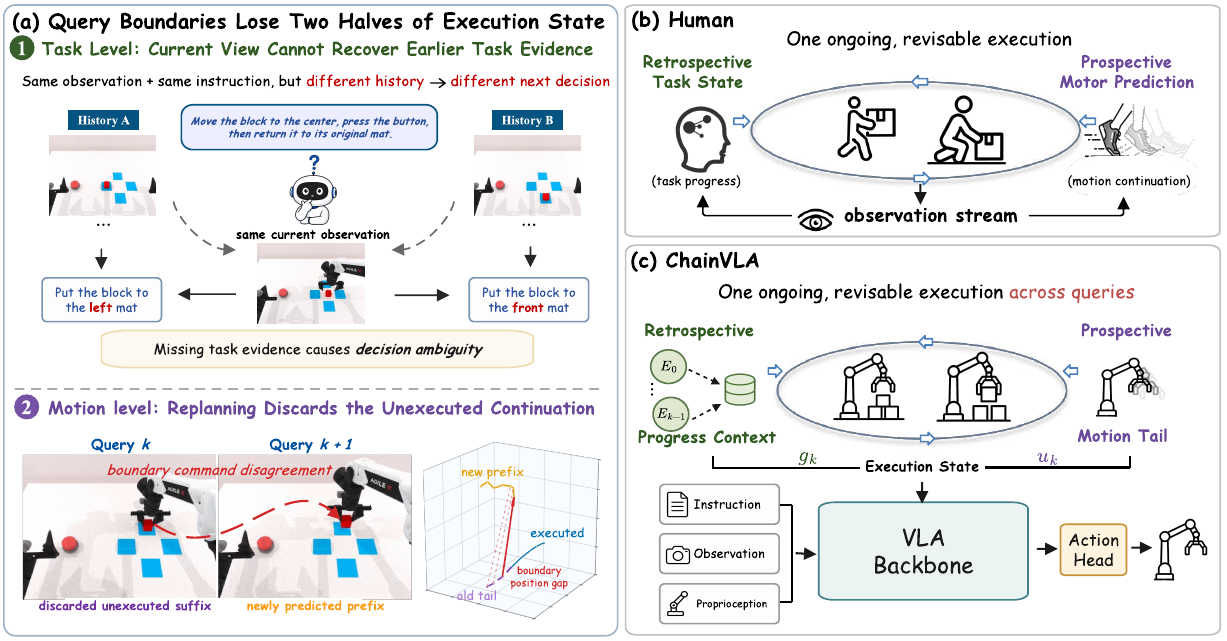}
    \caption{Motivation and design of ChainVLA. (a) At each query boundary, earlier task evidence may be absent from the current view, while replanning discards the preceding prediction's unexecuted continuation. (b) Human execution couples established task progress with a feedback-revised motor prediction, while executed motion shapes the observation stream used for subsequent progress estimation. (c) Inspired by this, ChainVLA carries both components across queries in a joint, revisable execution state comprising Progress Context $g_k$ and Motion Tail $u_k$, which together condition each new action horizon.}
    \label{fig:chainvla_teaser}
\end{figure*}

Long-horizon manipulation unfolds as a continuous execution in which each decision depends on both the progress established by earlier actions and the movement already underway.
When a person places several objects in their designated locations, deciding what to handle next depends on what has already been completed, while completing the current placement requires the ongoing movement to be maintained and revised.
These two requirements are supported by prior work on serial behavior~\cite{lashley1951serialorder,miller1960plans} and sensorimotor control~\cite{wolpert1995internalmodel}, and they correspond to task progress and motion continuation in Figure~\ref{fig:chainvla_teaser}(b).

In contrast to this continuous organization, action-chunked vision-language-action (VLA) policies execute long tasks through repeated receding-horizon queries, each of which predicts a finite action horizon, executes a strict prefix, and then replans~\cite{zhao2023act,chi2023diffusionpolicy,black2024pi0,black2025pi05,zheng2025xvla}.
Although consecutive queries are adjacent in time, each action chunk is generated anew from the current observation, proprioceptive state, and instruction, without an explicit handoff of either the task progress inferred at the preceding query or the unfinished continuation of its prediction.
Figure~\ref{fig:chainvla_teaser}(a) shows how this missing handoff creates ambiguity at two coupled levels.
At the task level, earlier task evidence may no longer be recoverable from the current view, as illustrated by RMBench's Put Back Block task, in which the same observation and instruction can follow different histories and therefore require different next decisions~\cite{chen2026rmbench}.
At the motion level, replanning discards the preceding prediction's unexecuted continuation, so the newly predicted prefix can disagree with the motion it replaces even when the current pose appears similar.
Because the motion executed after one query shapes the observation stream available at the next, motion continuity also helps preserve the observations needed to update task progress, suggesting that the two components operate consecutively rather than independently.

To recover the cross-query continuity lost by this design, existing attempts follow two complementary lines, with memory-augmented policies retaining recurrent state, selected observations, or compressed task history without preserving the preceding prediction's unexecuted continuation~\cite{shi2025memoryvla,chen2026keyframechaining,yang2026eventvla,huang2026robostream}.
A second line based on Temporal Ensemble, real-time execution, or native continuation preserves short-term motion by combining or reusing actions, yet it cannot retain task evidence from much earlier in the episode~\cite{zhao2023act,black2025realtimeexecutionactionchunking,liu2026learningnativecontinuationaction}.
This dilemma raises a central question: \textit{how can a VLA policy chain task progress and motion continuation across queries while keeping each new prediction responsive to the latest observation?}

\begin{figure*}[t]
    \centering
    \includegraphics[width=0.93\textwidth,pagebox=cropbox]
    {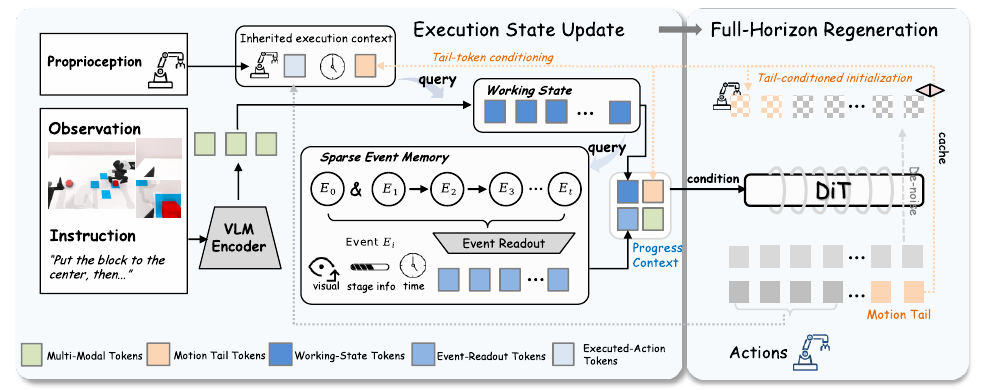}
    \caption{Overall query-transition architecture of ChainVLA. At query $k$, Progress Context combines live execution tokens from the recurrent Working State with event evidence retrieved from Sparse Event Memory. Motion Tail carries the preceding prediction's unexecuted suffix into state construction and action generation. The decoder regenerates the full horizon under the latest query input; prefix execution then advances task progress, and the newly predicted suffix becomes $u_{k+1}$ for the next query.}
    \label{fig:chainvla_transition}
\end{figure*}

To answer this question, we introduce \textbf{ChainVLA}, a 1.2B-parameter VLA policy that connects successive queries through a joint and revisable execution state (Figure~\ref{fig:chainvla_teaser}(c)).
The state comprises \emph{Progress Context}, which carries observation-derived task progress by combining a recurrent Working State with evidence retrieved from a sparse event memory, and \emph{Motion Tail}, which carries the preceding prediction's unexecuted continuation into state construction and action generation.
At each query, this joint state is read, updated, and passed forward, turning separately decoded action chunks into a cross-query chain.
Together, the two components condition each new action horizon, while the decoder regenerates every horizon position under the latest observation so that the carried state remains a revisable prior rather than a fixed plan (Figure~\ref{fig:chainvla_transition}).
With both components chained across queries, ChainVLA reaches 62.8\% average success on RMBench and 98.8\% across four LIBERO suites, whereas removing Motion Tail or Progress Context reduces RMBench success to 11.2\% and 3.0\%, respectively.


Based on this formulation and evaluation, our contributions are
\begin{itemize}
    \item We identify task progress and unfinished motion as two complementary components of cross-query execution state. Their asymmetric removal effects are consistent with motion continuity preserving the observation stream that informs task-progress inference.

    \item We introduce ChainVLA, a 1.2B-parameter policy that chains queries through a joint, revisable state of Progress Context and Motion Tail, regenerating each horizon under the latest available observation.

    \item We validate ChainVLA on RMBench and four LIBERO suites. Matched removals leave both partial states far below the complete model; observation history, temporal ensembling, and post-decoding smoothing also fall short.
\end{itemize}

\section{Related Work}

\noindent\textbf{Task progress under partial observability.}
Partially observable control represents interaction history through a belief state or another recursively updated statistic~\cite{astrom1965optimal,smallwood1973optimal,kaelbling1998planning}, and RMBench and RoboMME make the need explicit for manipulation, where an initial layout, hidden object, or completed subtask can stay relevant after it leaves the current observation~\cite{chen2026rmbench,dai2026robomme}.
Recent VLA systems preserve this observation-derived evidence through episodic, object-centric, prospective, or multi-scale memory~\cite{chung2025embodiedslotssm,hu2025pam,guo2026chameleon,torne2026memmultiscaleembodiedmemory,zeng2026helmharnessenhancedlonghorizonmemory}.
MemoryVLA reads and writes a perceptual-cognitive bank at every timestep, then consolidates adjacent entries at capacity~\cite{shi2025memoryvla}, whereas sparse alternatives select which observations become long-range records~\cite{chen2026keyframechaining,zeng2026kemo,yang2026eventvla}, and recurrent queries and bounded caches propagate local state between calls~\cite{li2026rememvla,sun2026tempofit,lei2026vpwem,tan2026memoact}.
All of these statistics are derived from observations, so none carries what the policy itself last intended.

\smallskip
\noindent\textbf{Action chunks across policy queries.}
Action-chunking policies predict short horizons and repeatedly replan~\cite{zhao2023act,chi2023diffusionpolicy}.
Temporal ensembling combines separately decoded overlapping horizons at execution time~\cite{zhao2023act}, guided test-time sampling improves agreement during decoding~\cite{liu2025bidirectionaldecodingimprovingaction,malhotra2025selfguidedactiondiffusion}, and real-time flow execution and native continuation reuse overlap during execution or generation~\cite{black2025realtimeexecutionactionchunking,liu2026learningnativecontinuationaction}.
These target local action coordination rather than evidence from much earlier in the episode, so what distinguishes ChainVLA is not prior-conditioned generation alone but combining prediction-derived continuation with observation-derived task evidence in one per-query state update.
Long-horizon planners and world models instead decompose subgoals or imagine latent state~\cite{liu2026goal2skill,shi2026memoryvlatemporalmodelingmemory,sun2026himemwamhierarchicalmemorygatedworld}, while ChainVLA chains calls to one horizon-based policy without requiring a separate planner.

\section{Method}

\subsection{Incomplete State Handoff}

ChainVLA considers history-dependent control in the strict-prefix receding-horizon regime, $1\leq h_{\mathrm{exec}}<H$.
At query $k$, the policy receives an observation $o_k$, proprioceptive state $r_k$, and language instruction $\ell$, grouped as $x_k=(o_k,r_k,\ell)$.
It predicts a horizon-$H$ action sequence $A_k=(a_{k,1},\ldots,a_{k,H})$ and executes the first $h_{\mathrm{exec}}$ actions before querying again~\cite{zhao2023act,chi2023diffusionpolicy}.
Thus, each new query observes the scene produced by the executed prefix while replacing a nonempty suffix of the preceding prediction.

Let $\mathcal{H}_k$ and $\mathcal{H}'_k$ denote two interaction histories before query $k$, with induced current inputs $x_k$ and $x'_k$.
The current input is decision-insufficient when identical inputs admit no shared optimal next executed prefix:
\begin{equation}
x_k=x'_k,\qquad
\mathcal{E}^*(\mathcal{H}_k)
\cap
\mathcal{E}^*(\mathcal{H}'_k)=\varnothing,
\end{equation}
where $\mathcal{E}^*(\mathcal{H})$ is the set of optimal length-$h_{\mathrm{exec}}$ prefixes.
This criterion identifies histories whose difference matters to the next committed control decision.
The task-scale case arises when an initial layout, completed subtask, or earlier displacement remains decision-relevant after leaving view~\cite{chen2026rmbench,dai2026robomme,chen2026keyframechaining}.

The current input also omits prediction-side context.
Because only a prefix of $A_{k-1}$ was executed, its remaining suffix specifies where the preceding query proposed to take the unfinished motion, which is neither completed behavior nor a fixed command, and may need revision once the new observation arrives.
A mismatch there extends beyond one boundary: a sharp change in the newly decoded command alters the executed state that contributes to $x_{k+1}$, giving repeated mismatches a route into later queries.

We use \emph{execution state} for the policy-side context that spans queries, distinct from the physical scene state in $x_k$, and write it as $s_k=(g_k,u_k)$.
Its retrospective half is Progress Context $g_k=\operatorname{Fuse}(L_k,Z_k^e)$, recomputed at each query by combining live execution tokens $L_k$ with evidence $Z_k^e$ retrieved from the carried Working State and Sparse Event Memory.
Its prospective half is Motion Tail $u_k$, the preceding prediction's unexecuted suffix, which crosses the boundary directly.
Chaining the queries means recursively maintaining this state while approximating history-conditioned generation,
\begin{equation}
\Pi_\theta(\cdot\mid x_k,s_k)
\approx
\Pi^*(\cdot\mid\mathcal{H}_k),
\qquad
s_{k+1}=F_\theta(s_k,x_k),
\end{equation}
where $\Pi_\theta$ is the learned policy, $\Pi^*$ an optimal history-conditioned policy over action horizons, and $F_\theta$ the query transition specified in Eq.~\ref{eq:transition}.
A belief state summarizes history through executed actions and observations, so it covers only $g_k$, whereas $u_k$ carries actions that have not occurred and thus has no counterpart there.
Both halves remain conditional inputs to a new prediction under $x_k$.

\subsection{The Execution State}

ChainVLA turns the incomplete handoff into one read, update, and write transition, while keeping every carried signal revisable under the newest observation.
Figure~\ref{fig:chainvla_transition} follows this transition from query $k$ to query $k+1$.
Query $k$ reads the current input $x_k$, recent executed actions $D_k$, previous Working State $W_{k-1}$, Sparse Event Memory $C_k$, and incoming Motion Tail $u_k$.
The encoder maps $x_k$ to multimodal tokens $X_k$.
Learned working queries then produce live tokens $L_k$ and an execution summary $\bar W_k$.
The summary retrieves event evidence $Z_k^e$ from $C_k$, and policy-facing fusion forms Progress Context $g_k=\operatorname{Fuse}(L_k,Z_k^e)$.
The incoming Motion Tail acts as a continuation prior through two model-side routes, where its tokens condition the Working State and remain in the decoder condition bundle, while its aligned trajectory initializes action generation.
The two halves differ in temporal coverage rather than update frequency, since Progress Context can carry evidence across many queries whereas Motion Tail connects only the immediately preceding prediction.

The conditional-flow decoder consumes $X_k$ and samples
$A_k\sim\Pi_\theta(\cdot\mid x_k,g_k,u_k)$ with a DiT backbone, so ChainVLA changes the decoder's cross-query context and initialization rather than the action-generator family.
The query transition is
\begin{equation}
\begin{aligned}
(A_k,W_k,C_{k+1})
&=F_\theta(x_k,D_k,W_{k-1},C_k,u_k),\\
u_{k+1}&=\operatorname{Tail}(A_k),
\end{aligned}
\label{eq:transition}
\end{equation}
where $F_\theta$ denotes the query transition and $\operatorname{Tail}(A_k)=(a_{k,h_{\mathrm{exec}}+1},\ldots,a_{k,H})$ denotes the predicted but unexecuted suffix.
This equation describes the transition shared by ordered training and rollout, and during training the carried suffix is detached before it conditions the next query.
After prefix execution, the controller appends the executed actions to form $D_{k+1}$ and commits the event candidate only when the dataset-specific write rule fires.
The next query reads $(D_{k+1},W_k,C_{k+1},u_{k+1})$ as its carried state.
The entire carried state is cleared between episodes.

\subsection{Carrying Task Progress}

Progress Context can carry task evidence across multiple queries, including evidence that remains decisive after it leaves the current view.
It combines a recurrent Working State for the current query with a retrievable Sparse Event Memory for sparse earlier milestones.

\paragraph{Recurrent Working State.}
The Working State summarizes the current execution and uses that summary to retrieve evidence from earlier queries.
Before query $k$, the recurrent state is $W_{k-1}=(L_{k-1},B_{k-1})$, where $L_{k-1}$ are the previous live tokens and $B_{k-1}$ is a bounded cache of earlier live tokens.
The update produces $W_k=(L_k,B_k)$ together with a pooled retrieval summary $\bar W_k$.

Let $\tau_k$ be the query-time encoding.
The executed history and incoming tail are encoded as $P_k=\phi_P(D_k)$ and $U_k=\phi_U(u_k)$.
Learned working queries attend to the current multimodal tokens and incorporate the preceding live state:
\begin{equation}
\begin{aligned}
\eta_k
&=\operatorname{MLP}
\left[\phi_r(r_k),\operatorname{Pool}(P_k),\operatorname{Pool}(U_k),\tau_k\right],\\
\widetilde L_k
&=\operatorname{CrossAttn}(Q_W+\eta_k,X_k),\\
L_k
&=\operatorname{LN}\!\left(
\widetilde L_k+m_{k-1}^{\mathrm{live}}
\operatorname{CrossAttn}(\widetilde L_k,L_{k-1})
\right),\\
\bar W_k&=\operatorname{Pool}([L_k,B_{k-1},P_k,U_k]),
\end{aligned}
\end{equation}
where $Q_W$ denotes the learned working-query bank, $\phi_r$ denotes the proprioception encoder, $\eta_k$ is the pooled conditioning vector, $\widetilde L_k$ is the current-input update before recurrence, $m_{k-1}^{\mathrm{live}}\in\{0,1\}$ is the recurrence-validity mask, and $\operatorname{LN}$ denotes layer normalization.
The mask is zero at episode start and one otherwise.
Terms separated by commas inside brackets are concatenated as tokens.
Pooled action summaries form the working-query context, while the full encoded sequences contribute to $\bar W_k$.
The bounded cache admits the most recent live tokens, evicts the oldest slot, and is cleared between episodes.
The live tokens $L_k$ enter the policy-facing fusion directly, whereas $\bar W_k$ drives event retrieval and, when used, stage estimation.
After retrieval, fusion of $L_k$ and $Z_k^e$ forms the policy-facing Progress Context.

\paragraph{Sparse Event Memory and Retrieval.}
Working State updates every query, whereas the Sparse Event Memory changes only on a triggered write.
At query $k$, it is the ordered sequence $C_k=\{(\kappa_i,E_i,b_i)\}_{i=1}^{N_k}$, where $N_k$ is the number of stored events.
The body $E_i=[V_i,S_i,T_i]$ stores learned visual features, Stage Info, and temporal features, while $b_i$ records the query at which the event was written for age encoding.
Learned projections of the current visual features, Working State summary, and query-time encoding produce the candidate $E_k=[V_k,S_k,T_k]$ and its retrieval key $\kappa_k$.

The first valid query is retained as an anchor, after which task-stage transitions determine later writes on stage-annotated data, whereas annotation-free data use a periodic query-time rule.
Retrieval is then conditioned on the current execution estimate rather than recency alone.
For stage-annotated data, the stage head yields $\pi_k=\operatorname{softmax}(h_s(\bar W_k))$ and the current Stage Info summary is $\bar s_k=\pi_k$.
Without stage annotations, $\bar s_k=\operatorname{Pool}(S_k)$.
The retrieval query is $q^e_k=\phi_q(\bar W_k,\bar s_k,\tau_k)$, and each record receives the age-aware score
\begin{equation}
\rho_{k,i}=\cos\!\left(
\operatorname{LN}(q_k^e),
\operatorname{LN}(\kappa_i+\operatorname{Emb}_{\mathrm{age}}(\operatorname{clip}(k-b_i)))
\right),
\end{equation}
where $\cos(\cdot,\cdot)$ denotes cosine similarity, $\operatorname{clip}$ bounds the event age to the supported range, and $\operatorname{Emb}_{\mathrm{age}}$ maps that age to an embedding.
The anchor and the highest-scoring non-anchor records are retrieved.
Cross-attention fuses their event bodies into readout tokens $Z_k^e$, which are then fused with $L_k$ to form Progress Context, and this read leaves the recurrent state unchanged until its next update.

\subsection{Carrying Unfinished Motion}

Motion Tail carries prediction-derived continuation from the immediately preceding query.
After prefix execution, ChainVLA caches the predicted but unexecuted suffix of $A_k$ as $u_{k+1}$.
This suffix gives the next query a reference for how the preceding motion was expected to continue.
Full-horizon execution leaves no unexecuted suffix but defers observation-conditioned replanning until the horizon ends.
Motion Tail instead operates in the strict-prefix regime, carrying the unfinished continuation while retaining frequent replanning.
We use \emph{motion intent} for this action-space continuation proposal, not for a separate symbolic plan or latent goal.
These actions have not occurred, since $D_k$ records executed behavior, whereas Motion Tail records the continuation proposed by the preceding decoder and imposes no fixed action on the next output.
Unlike post-decoding aggregation, Motion Tail is available while the next state and horizon are being formed, before any new output exists.

At query $k+1$, the incoming tail follows the two model-side routes shown in Figure~\ref{fig:chainvla_transition}.
Action-space normalization and a learned projection encode $u_{k+1}$ as tail tokens $U_{k+1}$ for the Working State.
For action generation, $\mathcal{I}$ aligns the normalized suffix with the full-horizon action representation.
The conditional initialization is
\begin{equation}
\begin{aligned}
\mu_{k+1}&=\mathcal{I}(u_{k+1}),\\
\epsilon_{k+1}&\sim\mathcal{N}(\mathbf{0},\mathbf{I}),\\
\widetilde A_{k+1}^{(0)}&=\mu_{k+1}+\sigma_u\epsilon_{k+1},
\end{aligned}
\end{equation}
where $\widetilde A_{k+1}^{(0)}$ denotes the initial full-horizon decoder state, $\mu_{k+1}$ is its tail-aligned mean, $\epsilon_{k+1}$ is standard Gaussian noise, $\sigma_u>0$ sets the noise scale, and $\mathbf{I}$ is the identity matrix.
Equivalently, $\widetilde A_{k+1}^{(0)}\mid u_{k+1}\sim\mathcal{N}(\mu_{k+1},\sigma_u^2\mathbf{I})$.
The conditional-flow decoder regenerates every horizon position from this state using the latest observation, Progress Context, and tail tokens.
Within the decoder, no horizon position is copied or frozen, so the current query may revise any action.
After decoding, the controller linearly blends overlapping steps of the regenerated plan and incoming suffix before execution.
This post-decoding operation is an execution-side handoff, separate from the two model-side routes.
At episode start, the tail path is masked, so the decoder uses its standard noisy initialization without a continuation prior.

\subsection{Training}

Training unrolls the same execution-state transition used at rollout and follows episode query order.
During this unroll, the next query may use the detached suffix predicted by its predecessor, and when that tail is not applied, the tail path is masked and the decoder uses its standard noisy initialization.
At every step, the model updates the Working State, retrieves Progress Context, predicts an action horizon, conditionally writes an event, and passes its detached Motion Tail forward to directly condition the next query.

When task-stage annotations are available, ground-truth transitions supervise the stage estimate and training-time write targets, although the labels never enter the action decoder.
At rollout, the predicted stage distribution supplies event semantics and triggers writes, whereas annotation-free domains use the same periodic query-time write rule during both training and rollout phases.

Training combines the standard conditional-flow action objective with stage supervision where annotations are available and an overlap-consistency regularizer for Motion Tail handoff.
The regularizer penalizes disagreement over the temporally aligned overlap of consecutive predicted horizons.
The preceding prediction is detached, encouraging a coherent handoff without fixing actions in the new horizon.
This training objective is distinct from the three-step overlap RMSE used only as a deployment diagnostic.
The recurrent and event states follow inference order, and rollout carries only the model-predicted suffix across the boundary.

\section{Experiments}

Our experiments ask three questions.
Does chaining execution state across queries support memory-dependent and general manipulation (Section~\ref{sec:main_results})?
What happens when each component is removed (Section~\ref{sec:ablations})?
And what is the relationship between motion continuity and memory-dependent task performance (Sections~\ref{sec:why_motion}--\ref{sec:not_smoothness})?

\subsection{Setup}

\paragraph{Benchmarks and comparisons.}
RMBench targets memory-dependent manipulation, where the current observation may be insufficient without earlier task-relevant states, and its five tasks require initial-state recall, target and order memory, return-location memory, or earlier-stage decisions~\cite{chen2026rmbench}.
We train one task-adapted model per task, evaluate 100 rollout episodes, and use RMBench as the primary diagnostic setting.
LIBERO covers the Spatial, Object, Goal, and Long suites and serves as a breadth check, with one model finetuned per suite and 50 episodes per task~\cite{liu2023libero}.
Comparisons cover action-chunk and pretrained policies together with benchmark and recent memory systems~\cite{chi2023diffusionpolicy,zhao2023act,black2024pi0,black2025pi05,zheng2025xvla,kim2024openvla,qu2025spatialvla,pertsch2025fast,nvidia2025gr00tn1,li2025cronusvla,chen2026rmbench,sridhar2025memer,shi2025memoryvla,tan2026memoact,hu2025pam,liu2026goal2skill}.

\paragraph{Protocol.}
All internally trained conditions share the Florence-2-large backbone~\cite{xiao2024florence2} and conditional-flow decoder, the same task data with 50 demonstrations per task, 20{,}000 optimization steps at batch size 4, and the same evaluation protocol, where Full has 1.2B parameters.
Queries stay in episode order during training, and every variant receives executed-action history separately from Motion Tail, so a removal withholds carried state rather than access to past actions.
The appendix specifies each intervention, the boundary-metric definitions, and the remaining training and rollout hyperparameters.

\begin{table*}[!t]
\centering
\small
\setlength{\tabcolsep}{2.0pt}
\begin{tabular}{@{}lrrrrrr@{}}
\toprule
\rowcolor{tablegroupgray}
\multicolumn{7}{c}{\textbf{RMBench}}\\
\midrule
Policy & Obs. & Rearr. & Put B. & Swap B. & Swap T. & Avg.\\
\midrule
DP & 1 & 0 & 0 & 11 & 20 & 6.4\\
ACT & 1 & 29 & 0 & 2 & 2 & 6.8\\
$\pi_{0.5}$ (2.6B+0.3B) & \underline{9} & 13 & 11 & 24 & 15 & 14.4\\
X-VLA (0.9B) & \underline{9} & 13 & 18 & 16 & 3 & 11.8\\
MemER & 7 & 17 & 0 & 14 & 7 & 9.0\\
MemoryVLA (7B+0.3B) & 0 & 22 & 50 & 17 & 9 & 19.6\\
Mem-0 (8B+2B) & 4 & 89 & \underline{90} & \underline{67} & 14 & \underline{52.8}\\
MemoAct & 4 & \textbf{98} & 41 & -- & \textbf{55} & --\\
PAM & -- & 29 & 30 & 15 & -- & --\\
Goal2Skill & 8 & 38 & -- & -- & -- & --\\
\midrule
\rowcolor{tablefullgray}
\textbf{ChainVLA (1.2B)} & \textbf{11} & \underline{93} & \textbf{96} & \textbf{74} & \underline{40} & \textbf{62.8}\\
\bottomrule
\end{tabular}
\hspace{0.5em}
\begin{tabular}{@{}lrrrrr@{}}
\toprule
\rowcolor{tablegroupgray}
\multicolumn{6}{c}{\textbf{LIBERO}}\\
\midrule
Policy & Spatial & Object & Goal & Long & Avg.\\
\midrule
OpenVLA (7B) & 84.7 & 88.4 & 79.2 & 53.7 & 76.5\\
SpatialVLA (3.5B) & 88.2 & 89.9 & 78.6 & 55.5 & 78.1\\
$\pi_0$-FAST (3B) & 96.4 & 96.8 & 88.6 & 60.2 & 85.5\\
GR00T-N1 (2.2B) & 94.4 & 97.6 & 93.0 & 90.6 & 93.9\\
$\pi_0$ (3.3B) & 96.8 & 98.8 & 95.8 & 85.2 & 94.2\\
$\pi_{0.5}$ + KI & 98.0 & 97.8 & 95.6 & 85.8 & 94.3\\
PAM & 86.1 & 97.0 & 94.5 & 84.7 & 90.6\\
MemoryVLA (7B+0.3B) & \textbf{98.4} & 98.4 & 96.4 & 93.4 & 96.7\\
CronusVLA (7B) & 97.3 & \textbf{99.6} & 96.9 & 94.0 & 97.0\\
X-VLA (0.9B) & \underline{98.2} & 98.6 & \underline{97.8} & \underline{97.6} & \underline{98.1}\\
\midrule
\rowcolor{tablefullgray}
\textbf{ChainVLA (1.2B)} & \textbf{98.4} & \underline{99.4} & \textbf{99.2} & \textbf{98.2} & \textbf{98.8}\\
\bottomrule
\end{tabular}
\caption{Success rates (\%) on RMBench and LIBERO. Comparison rows follow the cited references. Parenthetical parameter counts give the stated configuration size where directly available, the `+' symbol separates components reported individually, and missing counts are not inferred. Best and second-best listed values are bold and underlined. RMBench averages are omitted for methods without complete five-task coverage on that benchmark.}
\label{tab:main_results}
\end{table*}

\paragraph{Metrics.}
The primary metric is task success rate over 100 episodes per RMBench task and 50 per LIBERO task.
RMBench macro averages include only methods with complete five-task coverage, and LIBERO uses an unweighted four-suite average while retaining Long separately.

To inspect what happens at a query boundary, we adapt four lower-is-better diagnostics from prior work on action-chunk transitions~\cite{black2025realtimeexecutionactionchunking,liu2026learningnativecontinuationaction,zhan2026seam}, where position and orientation command discontinuity ($\mathrm{CD}_p$, $\mathrm{CD}_R$) measure the pose gap at replanning, boundary second difference ($\mathrm{B2}$) the command-velocity change across it, and three-step overlap RMSE ($\mathrm{RMSE}_3$) the disagreement between the preceding plan's unexecuted suffix and the next plan.
The first three read executed commands and the last compares plans, and none measures physical smoothness or accumulated state deviation.
We pair Full and w/o Motion Tail rollouts under matched scenes on Put Back Block (20 pairs), Rearrange Blocks (30), and Swap Blocks (30), and pool task means by pair count.
The appendix gives the formulas and all per-task values for these matched rollout conditions.

\subsection{Main Results}
\label{sec:main_results}

On RMBench, ChainVLA reaches 62.8 average success at 1.2B parameters, ahead of the strongest listed method with complete five-task coverage (Mem-0, 52.8, at 8B+2B), and it is best or second-best on every task (Table~\ref{tab:main_results}).
Methods without complete coverage are excluded from the average.
The gap is largest where a subtask must be repeated or reversed: Put Back Block requires returning a block to a pad that is no longer identifiable in the current view, and ChainVLA reaches 96 against 90 for Mem-0 and 50 for MemoryVLA.
On LIBERO the policy averages 98.8 with every suite above 98, including 98.2 on LIBERO-Long.
LIBERO is near ceiling for recent policies, so it establishes breadth rather than separating designs, and the analysis below therefore uses RMBench, where all conditions are trained and evaluated under the same matched protocol.

\begin{table}[!t]
\centering
\small
\setlength{\tabcolsep}{1.65pt}
\begin{tabular}{@{}lrrrrrr@{}}
\toprule
Configuration & Obs. & Rearr. & Put B. & Swap B. & Swap T. & Avg.\\
\midrule
\rowcolor{tablegroupgray}
\multicolumn{7}{@{}l}{\(\blacktriangledown\)\ \textit{Task-progress side} (motion continuation kept)}\\
w/o Stage Ann. & \underline{10} & \underline{84} & \underline{78} & \underline{61} & 34 & \underline{53.4}\\
w/o Live Tokens & \underline{10} & 78 & 72 & 48 & 28 & 47.2\\
w/o Event Readout & 9 & 65 & 25 & 38 & 24 & 32.2\\
w/o Progress Ctx. & 6 & 4 & 0 & 1 & 4 & 3.0\\
\addlinespace[2pt]
\rowcolor{tablegroupgray}
\multicolumn{7}{@{}l}{\(\blacktriangledown\)\ \textit{Motion-continuation side} (task progress kept)}\\
w/o Tail Tokens & 8 & 78 & \underline{78} & 58 & \underline{38} & 52.0\\
w/o Traj. Init. & 7 & 60 & 45 & 45 & 33 & 38.0\\
w/o Motion Tail & 1 & 24 & 0 & 2 & 29 & 11.2\\
\addlinespace[2pt]
\rowcolor{tablegroupgray}
\multicolumn{7}{@{}l}{\(\blacktriangledown\)\ \textit{No carried state, and a substitute}}\\
w/o Both & 4 & 0 & 0 & 0 & 4 & 1.6\\
FIFO Hist. + TE & 8 & 58 & 47 & 40 & 25 & 35.6\\
\midrule
\rowcolor{tablefullgray}
\textbf{ChainVLA (Full)} & \textbf{11} & \textbf{93} & \textbf{96} & \textbf{74} & \textbf{40} & \textbf{62.8}\\
\bottomrule
\end{tabular}
\caption{Ablations on RMBench (100 episodes per task). Best non-Full values are underlined. ``w/o Stage Ann.'' writes events every three queries instead. ``w/o Both'' keeps the current observation and executed actions. ``FIFO Hist.\ + TE'' replaces the carried state with fixed-length observation history and post-decoding temporal ensembling.}
\label{tab:component_ablation}
\end{table}

\subsection{Ablation Analysis}
\label{sec:ablations}

Table~\ref{tab:component_ablation} removes each half of the carried state while keeping the other, and neither partial state approaches the complete one, since task progress without motion continuation gives 11.2, motion continuation without task progress gives 3.0, and removing both gives 1.6, against 62.8 for Full.
Two halves scoring in the single digits to low teens do not add up to 62.8, so these are not two independent improvements over a working policy, and an incomplete state performs close to no state at all.
Because each removal withdraws a pathway's inputs, training term, and deployment handling together, the numbers measure system dependence rather than isolating the effect of any single mechanism.

Partial removals interpolate as expected.
Dropping either Progress Context input alone, live tokens at 47.2 and event readout at 32.2, costs far less than dropping both, and the event readout matters most on Put Back Block, which falls from 96 to 25, the task whose decisive evidence leaves the view earliest.
On the motion side, trajectory initialization is the more load-bearing route (38.0) than the tail tokens (52.0), but neither alone accounts for the fall to 11.2.
Fixed three-query writes still reach 53.4, so the chain does not depend on stage annotations to write useful events.

The substitute control is the informative one.
FIFO History + Temporal Ensemble supplies observation history before decoding and smooths overlapping predictions after it, which is how existing systems approximate both halves without carrying state.
It reaches 35.6, below Full on every task and 27.2 points below overall.
Past observations and agreement between consecutive outputs thus recover part of what the chain provides, but not the part requiring the previous prediction to condition the next one before it is generated.

\subsection{Motion Continuity and Memory}
\label{sec:why_motion}

\begin{figure}[!t]
    \centering
    \includegraphics[width=0.90\columnwidth]{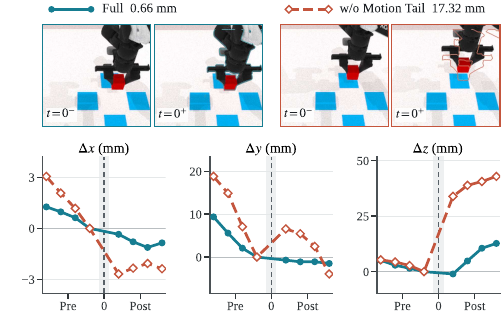}
    \caption{Illustrative Put Back Block boundary from paired-scene rollouts. Frames overlay the pre-boundary arm contour, and curves show right end-effector offsets from the $t=0$ anchor. This single case has 0.66/17.32~mm position CD for Full/w/o Motion Tail. Figure~\ref{fig:motion_tail_continuity} gives aggregate results.}
    \label{fig:motion_tail_boundary_example}
\end{figure}

\begin{figure}[!t]
    \centering
    \includegraphics[width=0.70\columnwidth]{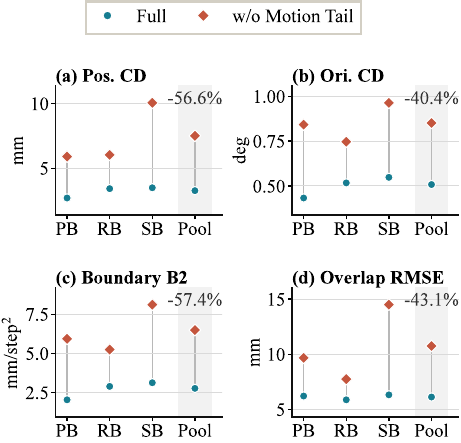}
    \caption{Boundary errors for Full and w/o Motion Tail. PB/RB/SB denote Put Back, Rearrange, and Swap Blocks, and Pool weights their 20/30/30 rollout pairs. Lower is better. Panels (a--c) use executed commands and (d) final plans.}
    \label{fig:motion_tail_continuity}
\end{figure}

RMBench is a memory benchmark, so the size of the motion-side effect needs explaining: removing motion continuation leaves the memory pathway fully intact, yet success falls from 62.8 to 11.2.
The likely reason is that the two halves act as consecutive links rather than parallel abilities.
When each horizon is sampled without reference to the one it replaces, consecutive commands can disagree at the boundary: on one matched Put Back Block boundary, the position gap is 17.32~mm without Motion Tail against 0.66~mm with it (Figure~\ref{fig:motion_tail_boundary_example}).
Across all 80 matched pairs, Full lowers every measure, with pooled reductions of 56.6\%, 40.4\%, 57.4\%, and 43.1\% for $\mathrm{CD}_p$, $\mathrm{CD}_R$, $\mathrm{B2}$, and $\mathrm{RMSE}_3$ (Figure~\ref{fig:motion_tail_continuity}).
A displacement of this size repeated at every boundary moves the arm along poses the demonstrations never visit, and recovering from them costs additional replanning queries.

That recovery is what reaches the memory pathway.
Progress Context is built from observations, so the sequence of viewpoints the arm visits is its input.
An episode spent recovering visits different viewpoints, in a different order, than the demonstrations that trained the retrieval, so evidence needed later may never be written or may be retrieved against an off-distribution query.
Motion continuity therefore appears to support reliable task-progress estimation rather than to act as a competing objective, which is also consistent with the asymmetry in Table~\ref{tab:component_ablation}.
Removing task progress leaves a policy that moves coherently toward the wrong subgoal and fails cleanly at 3.0, whereas removing motion continuation degrades the very input the surviving memory pathway reads.
Because the paired policies diverge within matched scenes, these boundary measurements establish an association rather than component-level causation.

\subsection{Post-Decoding Smoothing Controls}
\label{sec:not_smoothness}

\begin{table}[!t]
\centering
\small
\setlength{\tabcolsep}{3.4pt}
\begin{tabular}{@{}lrrrr@{}}
\toprule
& \multicolumn{2}{c}{Put Back Block} & \multicolumn{2}{c}{Swap Blocks}\\
\cmidrule(r){2-3}\cmidrule(l){4-5}
Method & Succ. & $\mathrm{CD}_p$ & Succ. & $\mathrm{CD}_p$\\
\midrule
w/o Motion Tail & 0 & 5.90 & 2 & 10.06\\
Linear Cont. & 0 & \textbf{2.09} & 2 & 3.72\\
Temporal Ens. & 0 & 3.57 & 0 & 7.52\\
\rowcolor{tablefullgray}
\textbf{Full} & \textbf{96} & 2.69 & \textbf{74} & \textbf{3.49}\\
\bottomrule
\end{tabular}
\caption{Post-decoding controls. Success uses 100 episodes per task, while position CD (mm, lower is better) uses 20 Put Back and 30 Swap pairs. The appendix reports the other boundary measures.}
\label{tab:boundary_extensions}
\end{table}

If the gain came from smoother boundaries, then producing smooth boundaries by any means should recover it.
It does not.
Linear Continuation reaches boundary values below Full on all four Put Back Block measures and three of four on Swap Blocks, yet succeeds in 0\% and 2\% of episodes against 96\% and 74\% for Full (Table~\ref{tab:boundary_extensions}).
Temporal Ensemble also reduces every measure relative to w/o Motion Tail and succeeds in 0\% of episodes on both tasks.

The comparison isolates when the suffix is used rather than whether it is used at all.
Full also blends overlapping steps after decoding, and Linear Continuation is exactly Full with that blend retained and the model-side conditioning removed, so the blend is held fixed and 96\% against 0\% is attributable to the pathway that differs.
Smoothing acts on already-decoded commands, so it can only interpolate between a plan and the motion it replaces, and the plan is fixed by the time it applies.
Carrying the suffix into state construction and generation instead changes what is decoded.
The boundary metrics are therefore diagnostics, not objectives: Full is not the smoothest condition here, yet the smoothest conditions still fail to recover task success.

\FloatBarrier
\section{Conclusion}

ChainVLA carries observation-derived task progress and the preceding prediction's revisable suffix across otherwise independent receding-horizon queries.
In our fixed-rate evaluation, neither partial state approaches the complete policy on RMBench, and their asymmetry is consistent with motion continuity preserving task-progress observations.

\bibliography{references}

\clearpage
\appendix
\renewcommand{\thesection}{\Alph{section}}
\setcounter{section}{0}
\setcounter{figure}{0}
\renewcommand{\thefigure}{A\arabic{figure}}
\setcounter{table}{0}
\renewcommand{\thetable}{A\arabic{table}}
\setcounter{equation}{0}
\renewcommand{\theequation}{A\arabic{equation}}

\section{Query Transition and Rollout Details}

At query $k$, ChainVLA reads the current multimodal input $x_k$, executed-action history $D_k$, preceding Working State $W_{k-1}$, Sparse Event Memory $C_k$, and incoming Motion Tail $u_k$.
It updates the Working State, retrieves task-relevant event evidence, and conditions a full-horizon prediction on Progress Context and Motion Tail.
After executing a prefix, the controller appends the executed actions to form $D_{k+1}$, may commit an event according to the dataset-specific write rule, and carries the predicted but unexecuted suffix forward as $u_{k+1}$:
\begin{equation}
\begin{aligned}
(A_k,W_k,C_{k+1})
&=F_\theta(x_k,D_k,W_{k-1},C_k,u_k),\\
u_{k+1}&=\operatorname{Tail}(A_k),
\end{aligned}
\end{equation}
where $F_\theta$ denotes the query transition and $\operatorname{Tail}(A_k)=(a_{k,h_{\mathrm{exec}}+1},\ldots,a_{k,H})$ is the unexecuted suffix.
This transition is shared by ordered training and rollout, with the carried suffix detached before it conditions the next query during training.
All carried states and validity masks reset between episodes, and Algorithm~\ref{alg:supp_rollout} states the induced loop.

\begin{table*}[t]
\centering
\footnotesize
\setlength{\tabcolsep}{4pt}
\begin{tabular}{@{}>{\raggedright\arraybackslash}p{0.15\textwidth}>{\raggedright\arraybackslash}p{0.47\textwidth}>{\raggedright\arraybackslash}p{0.32\textwidth}@{}}
\toprule
\rowcolor{tablegroupgray}
Component & Contents & Persistence\\
\midrule
Live tokens & 8 current plus 24 prior tokens in a three-slot FIFO & Recurrently updated every query\\
Executed-action tokens $P_k$ & 4 tokens encoded from the 8 most recent actions in $D_k$ & Re-encoded from $D_k$ at each query\\
Tail tokens $U_k$ & 4 tokens encoded from the incoming Motion Tail $u_k$ & Re-encoded from $u_k$ at each query; zeroed at episode start\\
Event record & 64 visual + 8 Stage Info + 8 temporal tokens, retrieval key, write-query index & Capacity 16, oldest non-anchor evicted\\
\bottomrule
\end{tabular}
\caption{Cross-query execution-state components at a glance.}
\label{tab:supp_state_tokens}
\end{table*}

\subsection{Working State}

Table~\ref{tab:supp_state_tokens} summarizes the cross-query execution-state components.
The Working State $W_k=(L_k,B_k)$ contains the current live tokens and their bounded FIFO cache, while executed-action tokens and tail tokens are constructed separately from $D_k$ and $u_k$.
Eight learned queries attend to the current multimodal tokens and recurrently update the live tokens, which enter Progress Context directly.
A pooled summary of the live, FIFO, executed-action, and tail tokens contributes to the event-retrieval query.

\subsection{Sparse Event Memory and Stage Info}

Each stored event comprises 64 visual, 8 Stage Info, and 8 temporal tokens, together with a retrieval key and write-query index.
Executed-action and tail tokens contribute to event construction through the retrieval summary $\bar W_k$ but are not stored as separate fields in the event record.
The first valid event is retained as an anchor.
On RMBench, stage transitions supervise training writes, while rollout writes require a predicted stage change with confidence at least 0.55; stage labels do not enter the action decoder.
Retrieval uses 32 clipped age buckets to read the anchor and the two highest-scoring non-anchor records.
Cross-attention fuses their event bodies into readout tokens without overwriting the Working State.

\subsection{Motion Tail and Ordered Training}

Motion Tail follows two model-side routes: the incoming suffix is encoded into tail tokens that condition Working State construction (Table~\ref{tab:supp_state_tokens}), while an aligned action-space representation initializes the full horizon with additive noise.
Motion intent denotes this continuation proposal rather than a symbolic plan or latent goal.
The decoder uses both routes but regenerates every position under the current observation and Progress Context.
After decoding, the controller linearly blends overlapping steps of the regenerated plan and incoming suffix before execution.
This execution-side handoff is separate from the two model-side routes, and the first query uses standard initialization instead.

Training processes queries in episode order.
During this unroll, each eligible query uses its predecessor's detached predicted tail with probability 0.5; otherwise, the tail path is masked and standard noisy initialization is used.
The overlap-consistency loss has weight 0.2 when Motion Tail is enabled and is distinct from deployment overlap RMSE.

\begin{algorithm}[t]
\footnotesize
\caption{ChainVLA rollout for one episode}
\label{alg:supp_rollout}
\begin{algorithmic}[1]
\State $W_{-1}\gets\varnothing$, $C_0\gets\varnothing$, $D_0\gets\varnothing$, $u_0\gets\varnothing$
\State $k\gets0$
\While{episode not terminated}
  \State observe $x_k=(o_k,r_k,\ell)$ and encode to $X_k$
  \State $W_k,\bar W_k\gets\textsc{UpdateWorking}(X_k,W_{k-1},D_k,u_k)$
  \State $\bar s_k\gets\textsc{StageSummary}(\bar W_k)$
  \State $Z_k^e\gets\textsc{Retrieve}(C_k,\bar W_k,\bar s_k,\tau_k)$
  \State $g_k\gets\operatorname{Fuse}(L_k,Z_k^e)$
  \State $A_k\sim\Pi_\theta(\cdot\mid x_k,g_k,u_k)$
    \Comment{every position regenerated}
  \State $A_k\gets\textsc{BlendOverlap}(A_k,u_k)$
    \Comment{execution-side handoff}
  \State execute prefix $(a_{k,1},\ldots,a_{k,h_{\mathrm{exec}}})$
  \State $D_{k+1}\gets\textsc{Append}(D_k,\text{executed prefix})$
  \State $C_{k+1}\gets\textsc{MaybeWriteEvent}(C_k,W_k)$
  \State $u_{k+1}\gets\operatorname{Tail}(A_k)$
    \Comment{carried as a revisable prior}
  \State $k\gets k+1$
\EndWhile
\end{algorithmic}
\end{algorithm}

\section{Evaluation and Intervention Details}

\paragraph{Common training and rollout protocol.}
All internally trained conditions share the protocol in Table~\ref{tab:supp_protocol} and the same evaluation protocol.
Every boundary leaves a six-step unexecuted suffix as Motion Tail.
Queries remain ordered, and every variant receives executed-action history separately from Motion Tail, so a removal withholds carried state rather than access to past actions.
RMBench stage labels supervise training where enabled and rollouts use predicted stages, while w/o Stage Annotation and annotation-free LIBERO write every three queries during training and rollout.

\begin{table}[t]
\centering
\footnotesize
\setlength{\tabcolsep}{3pt}
\begin{tabular}{@{}>{\raggedright\arraybackslash}p{0.46\columnwidth}>{\raggedright\arraybackslash}p{0.46\columnwidth}@{}}
\toprule
\rowcolor{tablegroupgray}
Item & Setting\\
\midrule
Vision-language backbone & Florence-2-large\\
Action decoder & Conditional flow\\
Demonstrations per task / steps & 50 / 20{,}000\\
Batch size & 4\\
Horizon $H$ / executed prefix $h_{\mathrm{exec}}$ & 30 / 24\\
Full parameters & 1.2B\\
Learning rates & Backbone $1\times10^{-5}$, decoder and memory $1\times10^{-4}$\\
Learning-rate schedule & Cosine annealing\\
Training hardware & 4$\times$ NVIDIA L20 GPUs\\
Random seed / framework & 42 (single seed) / PyTorch under Linux\\
\bottomrule
\end{tabular}
\caption{Shared training and rollout protocol for all internally trained conditions.}
\label{tab:supp_protocol}
\end{table}

\begin{figure*}[t]
\centering
\includegraphics[width=\textwidth]{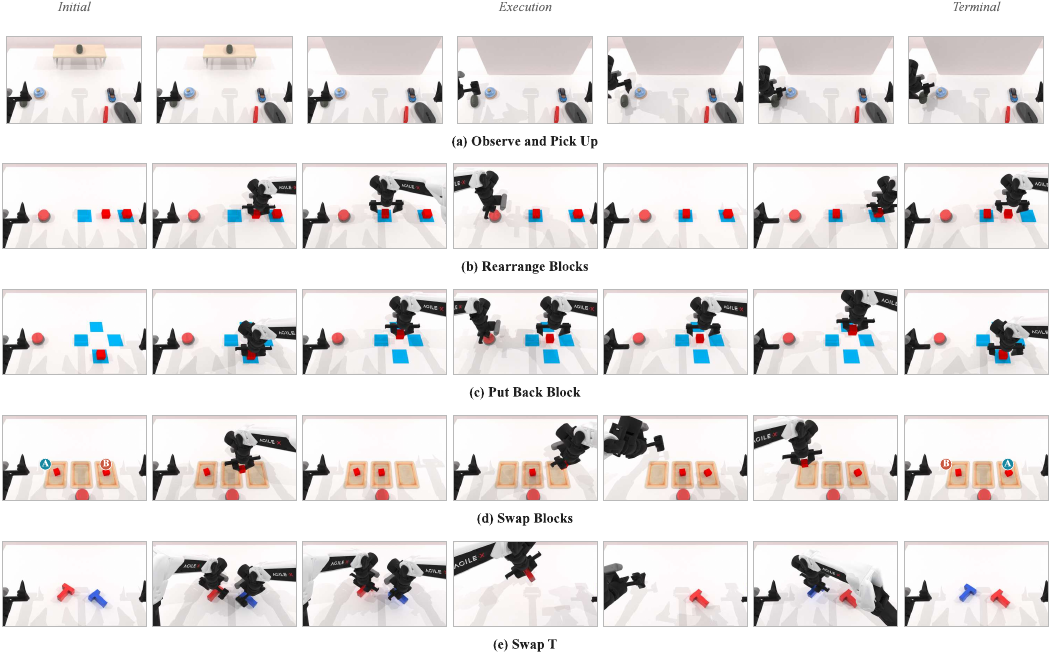}
\caption{RMBench task overview. Each row shows seven snapshots from one successful trajectory, ordered from the initial scene to the terminal state.}
\label{fig:supp_rmbench_tasks}
\end{figure*}

\paragraph{RMBench.}
The five RMBench tasks are bimanual and require several placements in sequence, so a single episode passes through multiple stages and therefore multiple query boundaries. Each internally evaluated condition uses 100 rollout episodes per task. Figure~\ref{fig:supp_rmbench_tasks} illustrates the intended terminal configuration for each task.

\paragraph{LIBERO.}
LIBERO serves as a breadth check over the Spatial, Object, Goal, and Long suites, with one model finetuned per suite. We evaluate all 10 tasks in each suite with 50 rollout episodes per task. The suites carry no stage labels, so writes follow the annotation-free schedule and the event chain grows on the fixed three-query interval in both training and rollout.

\paragraph{Optimization and infrastructure.}
Every condition trains from a single random seed with the learning rates, schedule, hardware, and framework listed in Table~\ref{tab:supp_protocol}.

\paragraph{Matched interventions.}
Matched controls cover all five RMBench tasks under the evaluation protocol above.
w/o Live Tokens retains the Working State for retrieval, w/o Event Readout removes retrieved evidence, and the Motion Tail controls remove either its Working State tokens or its trajectory initialization.
w/o Progress Context also removes its stage objective.
Table~\ref{tab:supp_ablation_definitions} gives the complete intervention semantics.

\begin{table}[t]
\centering
\small
\setlength{\tabcolsep}{2.5pt}
\begin{tabular}{@{}>{\raggedright\arraybackslash}p{0.32\columnwidth}>{\raggedright\arraybackslash}p{0.62\columnwidth}@{}}
\toprule
\rowcolor{tablegroupgray}
Configuration & Change from Full\\
\midrule
w/o Stage Annotation & Periodic writes; no stage supervision or stage-triggered writes\\
w/o Live Tokens & No live tokens in Progress Context; Working State still supports retrieval\\
w/o Event Readout & No retrieved event readout; recurrent live tokens retained\\
w/o Progress Context & No live tokens, event readout, or stage supervision; Motion Tail retained\\
w/o Tail Tokens & No tail tokens in Working State; tail-aligned initialization retained\\
w/o Trajectory Init. & Standard rather than tail-aligned initialization; tail tokens retained\\
w/o Motion Tail & No model-side tail routes, overlap training, or overlap blending; Progress Context retained\\
w/o Both & No Progress Context or Motion Tail; current observation, action history, and standard initialization retained\\
FIFO History + TE & FIFO observations, standard initialization, and Temporal Ensemble replace both carried contexts\\
\bottomrule
\end{tabular}
\caption{Matched interventions relative to Full. TE denotes Temporal Ensemble.}
\label{tab:supp_ablation_definitions}
\end{table}

\paragraph{Boundary rollouts and output-level controls.}
We pair Full and w/o Motion Tail rollouts under the same scenes on Put Back Block (20 pairs), Rearrange Blocks (30), and Swap Blocks (30), pooling task means by pair count.
On Put Back Block and Swap Blocks, Linear Continuation retains only the execution-side overlap blend without Motion Tail's model-side conditioning, while Temporal Ensemble aggregates decoded actions over time.
Their success rates use 100 episodes per task and method, and boundary means use the paired rollouts.
For compactness, the boundary tables use \emph{no-tail} for w/o Motion Tail.

\section{Boundary Diagnostics and Results}

\subsection{Diagnostic Definitions}

For a valid boundary $b$ and arm $a\in\{1,2\}$, let $(p_{b,a}^-,R_{b,a}^-)$ be the final executed Cartesian pose command before replanning, and let $(p_{b,j,a}^+,R_{b,j,a}^+)$ denote the command at post-boundary index $j$.
Averaging over the two arms, position CD, orientation CD, and translational boundary second difference are
\begin{equation}
\begin{aligned}
\mathrm{CD}_p(b)
&=\frac{1}{2}\sum_{a=1}^{2}
\lVert p_{b,0,a}^{+}-p_{b,a}^{-}\rVert_2,\\
\mathrm{CD}_R(b)
&=\frac{1}{2}\sum_{a=1}^{2}
\operatorname{angle}\!\left((R_{b,a}^{-})^{\top}R_{b,0,a}^{+}\right),\\
\mathrm{B2}(b)
&=\frac{1}{2}\sum_{a=1}^{2}
\left\lVert
(p_{b,1,a}^{+}-p_{b,0,a}^{+})
-(p_{b,0,a}^{+}-p_{b,a}^{-})
\right\rVert_2,
\end{aligned}
\end{equation}
where $\operatorname{angle}(R)$ denotes the $\mathrm{SO}(3)$ geodesic angle and the two finite differences in $\mathrm{B2}$ denote consecutive command increments.
We report $\mathrm{CD}_p$ and $\mathrm{B2}$ in millimeters and $\mathrm{CD}_R$ in degrees.
All three use the executed high-level command stream.

A boundary counts as valid when the preceding query contributed at least one executed command and the following query contributed at least two, since $\mathrm{B2}$ needs two consecutive post-boundary increments.
The first query of an episode has no predecessor and therefore contributes no boundary, and an episode that terminates immediately after replanning contributes none at that seam.
Both configurations are scored by the same rule on the same scenes, so the two arms of each pair see the same set of admissible seams up to where their rollouts diverge.

For consecutive final plans, let $p_{b,j,a}^{\mathrm{old}}$ and $p_{b,j,a}^{\mathrm{new}}$ denote the preceding suffix and new-plan positions over three aligned steps:
\begin{equation}
\mathrm{RMSE}_3(b)=
\sqrt{\frac{1}{6}\sum_{j=0}^{2}\sum_{a=1}^{2}
\left\lVert p_{b,j,a}^{\mathrm{old}}-p_{b,j,a}^{\mathrm{new}}\right\rVert_2^2},
\end{equation}
where $j\in\{0,1,2\}$ and $a\in\{1,2\}$ index the aligned steps and arms.
Boundary values are averaged within each rollout and then across rollouts within each task.
Let $c$ denote Full or no-tail (w/o Motion Tail). For metric $m$, the pooled mean and relative reduction are
\begin{equation}
\begin{aligned}
\bar m_{\mathrm{pool}}^{c}
&=\frac{20\bar m_{\mathrm{Put}}^{c}+30\bar m_{\mathrm{Rearr.}}^{c}+30\bar m_{\mathrm{Swap}}^{c}}{80},\\
\Delta_{\mathrm{rel}}
&=100\left(1-\frac{\bar m_{\mathrm{pool}}^{\text{Full}}}{\bar m_{\mathrm{pool}}^{\text{no-tail}}}\right)\%,
\end{aligned}
\end{equation}
where $\bar m_t^c$ denotes the rollout-averaged task mean and $\Delta_{\mathrm{rel}}$ denotes the pooled relative reduction from no-tail to Full.

\subsection{Full versus w/o Motion Tail}

\begin{table}[t]
\centering
\small
\setlength{\tabcolsep}{1.25pt}
\begin{tabular}{@{}llrrrr@{}}
\toprule
\rowcolor{tablegroupgray}
Task & Config. & $\mathrm{CD}_p$ & $\mathrm{CD}_R$ & $\mathrm{B2}$ & $\mathrm{RMSE}_3$\\
\midrule
Put Back & no-tail & 5.900 & 0.842 & 5.930 & 9.660\\
\rowcolor{tablegroupgray}
 & \textbf{Full} & 2.694 & 0.433 & 2.037 & 6.206\\
\addlinespace[1.5pt]
Rearrange & no-tail & 6.032 & 0.746 & 5.247 & 7.738\\
\rowcolor{tablegroupgray}
 & \textbf{Full} & 3.417 & 0.517 & 2.893 & 5.864\\
\addlinespace[1.5pt]
Swap Blocks & no-tail & 10.060 & 0.963 & 8.110 & 14.490\\
\rowcolor{tablegroupgray}
 & \textbf{Full} & 3.486 & 0.548 & 3.127 & 6.314\\
\addlinespace[1.5pt]
Pooled & no-tail & 7.510 & 0.851 & 6.491 & 10.751\\
\rowcolor{tablegroupgray}
 & \textbf{Full} & 3.262 & 0.508 & 2.767 & 6.118\\
\rowcolor{tablefullgray}
 & \textbf{Reduction (\%)} & \textbf{56.6} & \textbf{40.4} & \textbf{57.4} & \textbf{43.1}\\
\bottomrule
\end{tabular}
\caption{Full/no-tail boundary results over 20, 30, and 30 rollout pairs. Position CD, boundary second difference, and overlap RMSE use millimeters. Orientation CD uses degrees. Pooled means are count-weighted. Lower is better.}
\label{tab:supp_boundary_full}
\end{table}

Full is lower on all four diagnostics for every task in Table~\ref{tab:supp_boundary_full}.
After count-weighted pooling, the reductions are 56.6\%, 40.4\%, 57.4\%, and 43.1\% for position CD, orientation CD, boundary second difference, and overlap RMSE, respectively.

\subsection{Post-Decoding Controls}

Linear Continuation applies the execution-side overlap blend without Tail tokens or Tail-aligned initialization, while Temporal Ensemble aggregates actions over time.
These controls test whether reducing an output seam can replace Motion Tail's model-side conditioning.

\begin{table}[H]
\centering
\small
\setlength{\tabcolsep}{1.25pt}
\begin{tabular}{@{}lrrrrr@{}}
\toprule
\rowcolor{tablegroupgray}
Method & Succ. & $\mathrm{CD}_p$ & $\mathrm{CD}_R$ & $\mathrm{B2}$ & $\mathrm{RMSE}_3$\\
\midrule
\rowcolor{tablegroupgray}
\multicolumn{6}{@{}l}{\(\blacktriangledown\)\ \textit{Put Back Block}}\\
no-tail & 0 & 5.900 & 0.842 & 5.930 & 9.660\\
Linear Cont. & 0 & \textbf{2.090} & \textbf{0.391} & \textbf{1.660} & \textbf{5.170}\\
Temporal Ens. & 0 & 3.570 & 0.604 & 3.400 & 5.340\\
\rowcolor{tablefullgray}
\textbf{Full} & \textbf{96} & 2.694 & 0.433 & 2.037 & 6.206\\
\addlinespace[2pt]
\rowcolor{tablegroupgray}
\multicolumn{6}{@{}l}{\(\blacktriangledown\)\ \textit{Swap Blocks}}\\
no-tail & 2 & 10.060 & 0.963 & 8.110 & 14.490\\
Linear Cont. & 2 & 3.720 & \textbf{0.426} & \textbf{1.380} & \textbf{4.190}\\
Temporal Ens. & 0 & 7.520 & 0.731 & 6.510 & 8.210\\
\rowcolor{tablefullgray}
\textbf{Full} & \textbf{74} & \textbf{3.486} & 0.548 & 3.127 & 6.314\\
\bottomrule
\end{tabular}
\caption{Handoff controls. Success uses 100 episodes per task and method. Boundary means use 20 Put Back and 30 Swap Blocks pairs. Units follow Table~\ref{tab:supp_boundary_full}. Lower is better.}
\label{tab:supp_handoff_controls}
\end{table}

CD and B2 use executed commands; RMSE compares plans.
These metrics do not measure accumulated state deviation or fixed-frequency smoothness.
Linear Continuation has the lowest seam values but only 0--2\% success (Table~\ref{tab:supp_handoff_controls}).

\FloatBarrier
\section{Scope and Limitations}

\paragraph{Evaluation regime and statistical scope.}
All reported results come from fixed-rate simulation on RMBench and LIBERO, so the boundary diagnostics describe the commanded stream rather than physical trajectories under actuation limits.
We do not evaluate on hardware or vary the control rate, and the reported margins should not be read as predictions for asynchronous or real-time deployment.
Each internally trained condition uses a single random seed, so we report no seed variability or significance tests.
The reported success rates and boundary means should therefore be interpreted as point estimates under this protocol.

\paragraph{Interpretation of interventions and diagnostics.}
Each ablation withdraws a pathway from training and deployment together, so it measures dependence of the assembled system on that pathway rather than isolating an internal mechanism.
Likewise, Table~\ref{tab:supp_boundary_full} compares policies that diverge within matched scenes; it establishes an association between lower boundary diagnostics and higher success, not causation.
The asymmetry between the two state components is consistent with motion continuity protecting the observation sequence, but we do not claim a measured causal chain.
All four diagnostics are local to a seam and do not score whole-episode trajectory quality, accumulated state deviation, or fixed-frequency physical smoothness.
Success rate remains the criterion for task completion.

\paragraph{External baseline comparability.}
The source-reported comparison rows in the main paper were not rerun under our protocol and may differ in backbone, demonstration count, optimization budget, and success criterion.
Only the internally trained conditions share the settings listed above, so the ablation contrasts are matched while the external comparison is not.
We use the external rows for positioning and base the mechanism claims on the matched interventions.

\paragraph{Fixed design choices.}
Horizon and prefix lengths, the memory capacity of 16 records, the write-confidence threshold of 0.55, the fixed three-query write interval, and the overlap-consistency weight of 0.2 were each fixed once and reused across all conditions.
We did not search these values, so we cannot report sensitivity to them.

\end{document}